\documentclass{article}
\usepackage[final]{neurips_2020}
\usepackage{graphicx}
\usepackage{booktabs}       
\usepackage{hyperref}       
\usepackage{url}            
\usepackage{microtype}      

\title{Deep Neural Network Training without Multiplications}
\author{Tsuguo Mogami\\
Preferred Networks, Inc.,
Tokyo, Japan\\ 
\texttt{mogami@preferred.jp}
}
\date{October 2020}

\begin{document}

\maketitle

\begin{abstract}
    Is multiplication really necessary for deep neural networks? 
    Here we propose just adding two IEEE754 floating-point numbers with an integer-add instruction in place of a floating-point multiplication instruction. We show that ResNet can be trained using this operation with competitive classification accuracy. 
    Our proposal did not require any methods to solve instability and decrease in accuracy, which is common in low-precision training. 
    In some settings, we may obtain equal accuracy to the baseline FP32 result. This method will enable eliminating the multiplications in deep neural-network training and inference.
\end{abstract}

\section{Introduction}
Training deep neural networks is well known to be highly compute-intensive, with the multiplications typically being the most demanding in terms of the required hardware resources.
For this reason, various reduced-precision methods have been proposed for neural network training\citep{Dett, Gins, Mici, Cour}. 
The authors of \citep{Abad} and \citep{Dean} introduced BFLOAT16 format (1+8+7 format).
Despite these proposals, the multipliers still take the largest part of the circuit.

AdderNet \citep{Chen} reduced the number of multiplication by replacing the convolutions with $\ell_1$ norms, which only uses addition and absolute value, but it showed a mild ($\sim 1\%$) accuracy degradation. It is unclear how to generalize this method to layers other than convolution and fully-connected layers, which may limit applicable neural network architectures. It also required a special training procedure to make it stable.

The logarithmic number representation (LNR) is shown to give better accuracy than linear number representation with the same bit-width for deep neural networks \citep{miyashita2016convolutional, lee2017lognet}. These methods are mainly used for inference and have not been used for training of large networks and datasets such as ImageNet.
(Our proposal may be considered to be a natural generalization of LNR, which linearly interpolates it, and makes the training of large networks possible.)

To avoid hardware-intense multiplication, \citep{kim2018efficient} proposed using Mitchell's approximate multiplication \citep{mitchell1962computer} to convolutional neural networks and had shown only a slight ($\sim 0.2\%$) degradation in accuracy for VGG16, but the authors did not suggest using it for training.

In this article, we propose replacing the multiplication with integer addition (detail in Figure~\ref{list1}), and call it ‘addition-as-int’. 
We point out that adding two IEEE 754 floating-point representations with an integer addition instruction actually results in Mitchell's approximate multiplication.
We show that this operation may be used as a drop-in replacement of floating-point multiplication. We found only a small ($\sim 0.5\%$) or no accuracy degradation in training ResNet \citep{He} on the ImageNet dataset.

In addition to using our addition-as-int as an approximation to multiplication, we propose using it as a different but exact operation, which is expected to be more stable.  




\begin{figure}[t]
\begin{center}
\begin{tabular}{l}
\hline \\
{\tt float int\_mul(const float a, const float b)\{} \\
{\tt\hspace{2em} int c = *(int*)\&a + *(int*)\&b - 0x3f800000;} \\
{\tt\hspace{2em} return *(float*)\&c;} \\
{\tt\}} \\
\hline
\end{tabular}
\end{center}
\caption{\label{list1}
Our pseudo-multiplication in C++. This function gives a close ($\sim 12.5\%$) approximation of the FP32 multiplication. 
In detail, the sign bit is also computed correctly with this simple code if no underflow happens. We should supply rounding to zero if the underflow of the exponent is possible.
Exponent bias is 127 in IEEE754 representation. Then the exponent bias in 32bit integer representation is $127\times 2^{23} = {\tt 0x3F800000}$, since the LSB of the exponent is at the 23rd bit.}
\end{figure}


\section{Addition-as-int as float multiplication}
How can adding IEEE754 floating-point representations as mere integers (addition-as-int) give a close approximation to multiplication?
Intuitively, the reason is that it gives exact multiplication when either of the multiplicands is powers of 2. Otherwise, it will give linear interpolation between those exact values because the result is linear in the mantissa. As the result, the error is only 12.5\% in the worst case. ($1.5 \times 1.5$ gives $2$ in this operation, while $(1.5)^2$ is $2.25$.)

We may more easily understand our addition-as-int in terms of logarithms and exponentials. Let us consider two functions defined by
$e(x) =	1 + x$ for $0 =< x < 1$, otherwise
$e(x+1) = 2 e(x)$,
and
$l(x) = x - 1$ for $1 =< x < 2$, otherwise
$l(2x) = l(x) + 1$,
which give exponential of base 2 for integer $x$ and logarithm of base 2 when x is a power of 2 respectively, and give linearly interpolated values in between. The function $e(x)$ may easily be implemented reading an FP32 (IEEE754) representation into an integer register, subtracting the exponent bias, and considering the fixed decimal point to be at the left of the 22nd bit. The function $l(x)$ may easily be implemented by doing the same in the opposite direction. Then the addition-as-int as a substitute of multiplication of $a$ and $b$ may be obtained as $    f(a, b) = e(l(a) + l(b))$.

Backpropagation should be calculated using
$	\partial f(a, b) / \partial a = e’(l(a) + l(b)) l’(a),$
where the derivative $e’$ and $l’$ may easily be calculated by zeroing the mantissa. One drawback in implementing this operation is that when a forward path is $y = ab$ then backward gradient $da$ requires not only $b$ and $dy$ (like $da = b\ dy$) but also $a$, which results in an increase of bandwidth about $a$. 
The ``a-operations", which will be introduced later, does not have this drawback.

In the light of logarithm and exponential, a linear layer of neural networks 
$	y_i = \sum_j w_{ij} x_j$
may be considered as
$	y_i = \sum_j \exp( \log w_{ij} + \log x_j ),$
which is a log activation layer before addition before an exp activation layer.  Therefore replacing exp and log with our $e$ and $l$ function is just like changing the activation functions a little, and we may expect that the change of the result should be as little as replacing the ReLu activation function with another activation function.


\section{Experimental settings}
We used Chainer \citep{Toku} and CuPy \citep{Okut} to implement the layers and Nvidia PASCAL P100 GPUs to run the experiments.
The ResNet-50 was trained on ImageNet (ILSVRC 2015) dataset for 100-epoch as our benchmark. The standard augmentation methods, i.e. flip, crop, resize, were applied as \citep{He}. The training hyperparameters are lr=0.05*32/256, batchsize=32, the learning rate was decreased at 30, 60, and 90 epoch by the factor of 10, weight decay rate was $10^{-4}$, and momentum SGD with $\mu=0.9$, which are standard. All the FP32 baseline and other training are done under the same random seed.
%

In the MNIST training, a MLP of linear(784, 1000)-ReLu-linear(1000,1000)-ReLu-linear(1000,10) structure with an ordinary softmax-cross entropy layer was used without augmentation with Adam for 20 epochs with batchsize=100.

In implementing softmax cross-entropy, we have to be careful about the shift-invariance of the logsumexp function ${\rm lse}(\vec x)$ in it.
We have relationship ${\rm lse}(\vec x + a) = {\rm lse}(\vec x) + a$, and then we usually subtract the maximum element of $\vec x$ from $x_i$ before the calculation and add it afterward. In our approximate exponential function, this invariance holds only for integer $a$, and the implementation should respect that. We observed performance degradation without this consideration.

\begin{table}[t]
\caption{Top-1 accuracy (\%) on ImageNet of ResNet-50 with e-operations, a-operations.
The meaning of the abbreviation like ``cE" is explained at the beginning of the result section.}
\label{table1}
\begin{center}
\begin{tabular}{llll}
\toprule
\multicolumn{2}{c}{\bf e-operations}  &\multicolumn{2}{c}{\bf a-operations} \\
\cmidrule(r){1-2} \cmidrule(r){3-4}
baseline(FP32)  & 75.6 & & \\
cE  & 75.1 & ca.fE  & 75.6\\
cE.fE  & 75.0 & cE.fa  & fail\\
cE.fE.bE  & 73.6 & ca.fE.ba  & 75.5\\
cE.fE.bE.eE  & 73.9 & ca.fE.ba.ea  & 75.6\\
\bottomrule
\end{tabular}
\end{center}
\end{table}

\section{Results}
We present results for two variants of our addition-as-in (Figure~\ref{list1}). The first is the exact operation (``e-operations") scheme  and the second is the approximate operation (``a-operations") scheme.
Also, the results of using BFLOAT16 together will be presented.

We denote the network configuration as follows: `cE' denotes using e-operations for all the convolution layers, `ca' denotes using a-operations for all the convolution layers, and ordinary multiplication is used if none is specified. Other than `c' for convolution layers, `f' denotes a fully-connected layer, `b' for batch normalization layers, and `e' for a softmax cross-entropy layer. Thus `cE.ba' denotes that the network has convolution layers with e-operations, batch normalization layers with a-operations, and other layers with ordinary definition, for example.

\subsection{Exact operations}

Here we use our addition-as-int instead of multiplication and consider the exact differentiation of this function. We call such operations by ``e-operations", and each of the operations will be called ``e-mult", ``e-exp", ``e-div", and so on.
This e-mult is not approximating multiplication but replacing multiplication with a similar but different operation. 

For the MNIST dataset, applying e-mult to the simple MLP gave the accuracy of 0.980 and 0.982 for two runs of different initialization, where the ordinary multiplication gave 0.981 and 0.981, which may be considered to be the same.

Then we began ImageNet training with replacing the multiplications with our e-mult in all the convolution layers. We saw only a little ($\sim 0.5\%$) degradation in classification accuracy (cE in Table~\ref{table1}). Further replacing the multiplications in the fully-connected layer did not change the test accuracy (cE.fE in Table~\ref{table1}) from the cE case. 

\subsection{Approximate operations}

On the other hand, we may also use ${\rm int\_mul}(x, y) + \gamma$ as an approximation to multiplication, where the correction constant $\gamma$ will be explained shortly. The gradient of multiplication is another multiplication, and it can again be approximated with ${\rm int\_mul}(x, y)+\gamma$.
We call this use of operations as “a-operations”, and each of the operations will be called as “a-mult”, “a-exp”, “a-div”, and so on.

The constant $\gamma$ was determined as follows. If we want to use $l(x)$ as an approximation to $\log_2(x)$, we need to use $l(x)+\gamma$  because $l(x)$ is consistently below $\log_2(x)$. Likewise, the same applies to $e(x+\gamma)$ as an approximation to $2^x$.  We assume the distribution of the incoming numbers are flat and the average difference of $\log_2(x)$ and $l(x)$ should be canceled, so we have
$	\gamma = \int_1^2 (\log x - (x-1)) dx = 3/2 - 1/\log 2.$
We found this assumption to be sufficient for training neural networks.

We first replaced multiplications of the convolution layers with a-mults but kept e-mult for the fully-connected layer of ResNet-50. We observed equal accuracy (ca.fE in Table~\ref{table1}) in reference to the baseline FP32 model, which is better than e-operations. Replacing with a-mult for the fully-connected layer decreased accuracy more than 30\% at 10th epoch and no further training was performed. From this, we considered using a-operation in fully-connected layers is not good.

\subsection{Replacing all the other layers with our operations}

In addition to convolution and fully-connected layers, we replaced operations in batch normalization with e-operations. 
We saw mild accuracy degradation ($\sim 1.4\%$) without other trouble (cE.fE.bE in Table~\ref{table1}). We further replaced the softmax cross-entropy layer using e-log, e-exp and observed no further drop of accuracy (cE.fE.bE.eE in Table~\ref{table1}).
The experimental result for a-operations gave the same accuracy as FP32. We replaced the batch normalization using a-mult to find no difference (ca.fE.ba in Table~\ref{table1}) from FP32. Additionally, we replaced the softmax cross-entropy layer using a-operations to find equal accuracy (ca.fE in Table~\ref{table1}) as FP32.
Therefore, if one is considering new hardware that lacks multipliers, a-operation seems better for batch normalization.

\begin{figure}[t]
\begin{center}
\includegraphics[width=0.5\linewidth]{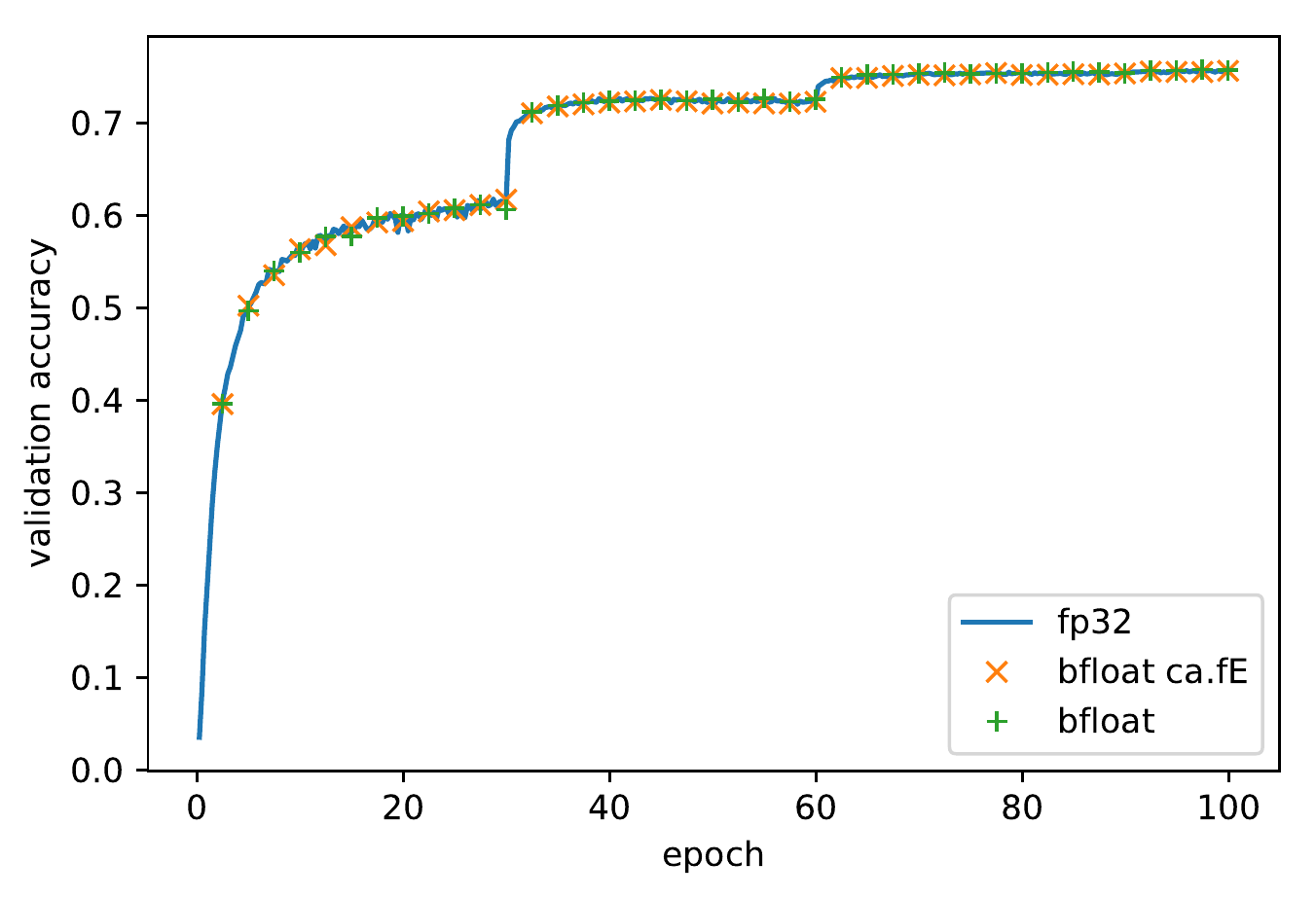}
\end{center}
\caption{\label{bfloat-lc}
Learning curves using BFLOAT16 for ImageNet.
We see no deviation from the FP32 result.
``ca.fE" means convolution layers with a-mult and a fully connected layer with e-mult.
}
\end{figure}

\subsection{Compatibility with a 16-bit float format}

Is it possible for our operation to cope with reduced-bit representation such as BFLOAT16? 
If it is successful we would also obtain the merits of both ours and BFLOAT16's. Methodologically, we did almost the same as \citep{Kala}, i.e. pseudo multiplications were done for and resulted in BFLOAT16, then they are accumulated in FP32, and weights were kept in FP32. Our method was only different in that weight gradients were rounded after accumulation before adding to the master weights.

As the result of ImageNet training, our BFLOAT16 baseline implementation did show the same accuracy (75.7\%) as the FP32 baseline, and we saw equal accuracy (75.7\%) for our approximate operations (ca.fE scheme) compared from the FP32 baseline (Figure~\ref{bfloat-lc}).
Here we showed that our method works with BFLOAT16 without any accuracy degradation.

\section{Conclusion and discussion}
In this article, we showed that replacing floating-point multiplications with integer additions gives competitive or sometimes equal performances as the FP32 baseline in training ResNet-50. We showed that all the multiplications in all the layers of the neural network can be replaced with integer additions. The experimental results also suggest that we may freely combine layers with our addition-as-int and ordinary layers. In addition, we showed that our method can work with a shorter floating-point format, i.e. BFLOAT16.



The future hardware that has fused instruction for our operation will run faster, though our integer-add-float-add operation ran slower for the current GPU, which is highly optimized for the fused-multiply-add.

Our results raise the possibility of further research on optimal activation functions. The multiplication may be considered as a combination of logarithmic activation function and exponential activation function. We have suggested only one way of changing these activation functions. 
It is then interesting to consider future research directions that focus on finding the most hardware efficient activation functions that also attain the highest accuracy.

\begin{ack}
The author thanks Brian Vogel and Kentaro Minami for reading the manuscript and their suggestions, and Tanvir Ahmed for comments.
\end{ack}

\section*{Broader impact}
The impact of this study is expected to make deep neural-network training more efficient.
No one would get any disadvantage out of this study, but every person training neural networks will potentially benefit from improved efficiency, and every person using neural network inference can also benefit.
No failure modes nor leverage of bias specific to this method is expected to exist since it works equivalently to the existing methods as far as the authors know.

\bibliographystyle{plain}
\bibliography{main}

\end{document}